\documentclass[sn-mathphys-ay]{sn-jnl}

\usepackage{chngcntr}
\usepackage{subcaption}
\usepackage{tcolorbox}
\usepackage{bm}
\usepackage{graphicx}%
\usepackage{sidecap}%
\usepackage{multirow}%
\usepackage{amsmath,amssymb,amsfonts}%
\usepackage{amsthm}%
\usepackage{mathrsfs}%
\usepackage[title]{appendix}%
\usepackage{xcolor}%
\usepackage{textcomp}%
\usepackage{manyfoot}%
\usepackage{booktabs}%
\usepackage{algorithm}%
\usepackage{algorithmicx}%
\usepackage{algpseudocode}%
\usepackage{listings}%
\usepackage{enumitem}
\usepackage{diagbox}
\usepackage{xcolor}
\usepackage{subcaption}

\counterwithout{figure}{section}
\counterwithout{table}{section}
\begin{document}

\title{AugAbEx : Bridging Abstractive and Extractive Legal Summarization}

\author{\fnm{Purnima} \sur{Bindal}}\email{pbindal@cs.du.ac.in}

\author{\fnm{Vikas} \sur{Kumar}}\email{vikas@cs.du.ac.in}

\author*{\fnm{Vasudha} \sur{Bhatnagar}}\email{vbhatnagar@cs.du.ac.in}


\affil{\orgdiv{Department of Computer Science}, \orgname{University of Delhi}, \orgaddress{\state{Delhi}, \country{India}}}







\abstract{Automatic summarization of legal judgments liberates law professionals from  heavy cognitive burden due to the complexity of the language, context-sensitive legal jargon, and the length of the document. Caveats of abstractive summarization for legal documents, revealed in recent studies, have impelled the development of new hybrid and extractive summarization methods, along with datasets in the legal domain.

We propose an efficient and elegant pipeline (AugAbEx) to repurpose an existing large case-summarization dataset  with  human-written gold-standard summaries by augmenting it with silver standard extractive summaries ensconcing  experts' opinion.   Availability of silver summaries along with gold standard human-written summaries will bolster development and evaluation of novel hybrid and extractive case-summarization algorithms in the  legal  domain. We  thoroughly scrutinize the augmented extractive  summaries  in structural, lexical, and semantic dimensions,  within a domain-specific framework, to  ensure  quality. Extensive experiments and statistical test on seven legal case-summarization datasets demonstrate that the  silver standard extractive summaries produced by the proposed pipeline score well  across all evaluation dimensions. Comparison of AugAbEx with two baselines and three current state-of-the-art methods reveals that it  outperforms all competing methods, and the quality of the summaries  produced by the proposed pipeline is  superior. 
}


\maketitle

\section{Introduction}
\label{sec1}
The importance and utility of case summaries for judgment writing, case preparation, and  simply explaining the case to a layperson  have garnered huge interest in legal case summarization research \citep{2021-where-are-we-survey-jain,2025-survey-comprehensive,2025-nlp-legal-domain-survey,2026-He-survey}. Legal case documents,  characterized by intricate  legal jargon, statutes, provisions, precedents, etc.,  pose  serious challenges due to the need to preserve legal terminology in the summaries.  The insistent  demand for  semantic and factual precision  in  case summaries further exacerbates these challenges.  

Though the ability of transformer-based approaches  to produce\textit{ human-like} summaries  has pushed development of  \textit{abstractive} summarization techniques in general, summarization of legal case documents using these methods is still unsatisfactory \citep{deroy2023ready,24-applicability-of-llm}. Despite  fluency  and \textit{human-like} characteristics of  abstractive summaries of case documents,  factual inaccuracies, paraphrase errors and misrepresentation of legal jargon in these summaries renders them  unreliable for legal professionals \citep{2023-llm-challenges}. \textit{Extractive} case summaries, on the other hand,  have been shown to better preserve the accuracy and language of the original case documents than their abstractive counterparts \citep{19-comparativeStudySummAlgo,22-ExtandAbsMethods,23-impact-of-nlp-law}. In practice,  legal experts are more concerned with getting the \textit{factually correct} information than fluidity in language \citep{24-cannot-be-right}. The law experts can pad the semantic gaps that arise due to lack of fluency and cohesion in extractive summaries, as long as concepts and the core legal entities are present in the summary. In a recent survey, \citet{2025-survey-comprehensive} find that despite lack of coherence and poor cohesion, more \textit{extractive} approach papers were published for legal case summarization during 2020 and 2024 than \textit{abstractive} method papers. 

Despite the pertinence of \textit{extractive} summarization of legal case documents, availability of legal summarization datasets with extractive gold standard summaries is severely limited. In a recent survey, \citet{2026-He-survey} list $18$ summarization datasets, of which only one has gold standard extractive summaries and the rest have  reference summaries  based on catchphrases, headnotes or abstractive annotations.  The scarcity of extractive gold standard reference summaries poses hurdle for  development and evaluation of novel summarization models in three ways. \textit{First,} it retards the development of hybrid case summarization approaches, which are contingent on extractive reference summaries to effectively handle long documents and improve abstractive summaries. \textit{Second,} it impedes the development of supervised approaches for extractive summarization, which rely on the availability of extractive gold-standard summaries to assign sentence-level labels to the judicial text.  \textit{Third,}  the evaluation of extractive summarizers using abstractive reference summaries underestimates their performance because lexical overlap between them may be low even when the extractive summary is semantically correct \citep{2024-rouge-limitation-recsumm, 2024-hybrid-DCESumm,2025-survey-comprehensive}. 
\subsection{Related Works}
Recent surge in hybrid approaches for legal case summarization has highlighted the need for extractive reference summaries for generating abstractive summaries \citep{2023-two-stage,2025-hybrid-with-domain-knowledge,2026-He-survey,2026-hybrid-billsum,2026-optimized-hybrid-llama2}. Limited availability of extractive reference summaries has been circumvented  by including an additional step in the pipeline to  construct  an extractive case summary consisting of the key sentences rich in legal terminology.   Several innovative mechanisms have been employed to pick salient sentences  from the judicial text, including  BERT+LSTM model \citep{2023-two-stage},  LLM-assisted annotation pipeline \citep{2025-comp-pipeline-LTS},  TextRank-based top-k sentence selection \citep{2026-optimized-hybrid-llama2, 2004-textrank}, BillSum-guided enhancement \citep{2026-hybrid-billsum}. An alignment framework for matching judgment chunks with gold-standard summary sentences has been introduced in prior work \citep{2022-oag-moro-semantic,2023-hybrid-align-moro}. Next, the aligned chunks are then transformed to abstractive case summary using either a deep neural method or an LLM.

Consistent and precise legal language in the gold standard summaries is essential for training high-fidelity case summarizers. Development of neural classifiers and sentence ranking models for extractive summarization is contingent on availability of binary sentence labels to indicate their inclusion or exclusion in summary. Paucity of extractive reference summaries for development of supervised extractive summarizers has been skirted by leveraging ROUGE-based alignment between the abstractive summary sentences and original document sentences   \citep{2023-soft-labeling-graph-moro,2024-moro-multilanguage, 2024-soft-labeling-cross-ragazzi, 2026-soft-labeling-sende}. \citet{2024-hybrid-DCESumm} utilize LegalBERT-based contextual representation of sentences for alignment and then construct summary using unsupervised clustering based method. Other approaches for crafting abstractive summaries include  graph-neural network-based method \citep{2024-soft-labeling-cross-ragazzi}, CNN-based model \citep{2026-soft-labeling-sende}, LLM \citep{2021-labeling-parikh-lawsum,2024-rouge-limitation-recsumm}.

Since legal summarization  datasets with extractive reference summaries serve as a strong and reliable foundation for training and evaluating the performance of extractive case summarizers, creation of high-quality extractive summary datasets is imperative for the progress in extractive case summarization research.  However, crafting human-annotated extractive summaries  is a humongous task due to the sheer volume of cases, time constraints and cognitive burden on law professionals. Automatic generation of extractive summaries   is a prudent  and practical  alternative. Integrated in the pipeline and termed as \textit{silver standard reference summaries}, such summaries have been used as a substitute, approximation, or auxiliary alternative to a human-written reference summary \citep{2017-silver-johnson,24-silver-summaries}.
\subsection{Proposed Solution}
In order to automatically generate \textit{silver standard reference summaries} for augmenting an abstractive  case summarization dataset, we  leverage the human-written abstractive gold standard summaries in the existing case summarization datasets. We transform the original abstractive summaries (OAG) to their corresponding extractive counterparts to serve as \textit{silver} summaries in research, development and evaluation of  hybrid and extractive summarizers. The proposed solution  not only re-purposes existing abstractive case summarization datasets, but can also be used for creating  large extractive summarization datasets from millions of cases curated by prevalent legal case repositories globally\footnote{\begin{tabular}{l} 
\url{https://legal.thomsonreuters.com/en/westlaw} \\
\url{https://www.liiofindia.org/} \\
\url{http://www.loc.gov/crsinfo/} \\
\url{https://www.supremecourt.uk/decided-cases/} \\
\url{https://primelegal.in/blog/}\\ 
\url{https://lawtimesjournal.in/category/case-summary/}
\end{tabular}} that maintain human-written summaries as \textit{head-notes}. 

The objective is achieved by engineering a light, transparent and cost-effective pipeline inspired by several existing works. Furthermore, re-purposing human-written abstractive summaries begets an additional advantage of strong alignment of the transformed extractive summaries with the expert's opinion about the important aspects of the case at a meager cost. Ensuring the quality of extractive silver summaries is de rigueur to instill confidence in the the repurposed datasets.  We perform a thorough comparative analysis of \textit{original abstractive gold} reference (OAG) summaries  written by law experts and the \textit{transformed extractive silver standard} (TES) summaries along  structural, lexical, semantic and domain-specific dimensions. We also  scrutinize the prevalence of legal entities in the two summaries to compare their relative richness in legal terminology, which is instructive for law practitioners. We further reinforce our investigation by performing a pairwise comparison of the OAG  and TES summaries using the Bradley-Terry model \citep{21-bt-test} for the four dimensions. 

 \noindent\textit{Contributions and Organization of the paper:}
Achieving our overarching objective to \textit{re-purpose} abstractive case summarization datasets to support further research and developments in the area of extractive case summarization, we list below  our specific contributions.  
\begin{enumerate}[label=\roman*)]
    \item We propose AugAbEx pipeline to transform original abstractive gold (OAG) summaries to corresponding extractive silver (TES) summaries, while preserving the expert's assessment of the salient ideas in the case documents (Section \ref{sec:method}).
    \item We propose a comprehensive framework for evaluating the quality of transformed summaries for structural, lexical, and semantic dimensions. We also compare the quantum of legal information present in the OAG and TES summaries. Additionally, we perform a thorough pairwise comparison of the OAG and TES summaries using Bradley-Terry test, a robust statistical approach for comparative analysis of two systems (Section \ref{sec:eval-metric}).
     \item We transform seven popular English case summarization datasets with original abstractive gold (OAG) summaries to corresponding extractive silver (TES) summaries. The augmented datasets\footnote{The transformed extractive silver standard (TES) summaries will be released after the manuscript review process is completed.} are a valuable resource for the development of much-needed extractive summarization algorithms in the legal domain (Section \ref{sec:datasets}).
     \item We evaluate the TES summaries for the augmented datasets across four dimensions using multiple evaluation attributes. Additionally, we benchmark the quality of TES summaries against summaries produced by several baselines and state-of-the-art legal summarization methods, and the results underscore the superiority of the proposed pipeline (Section \ref{sec:expt}).
\end{enumerate}
\section{AugAbEx: Pipeline for Augmenting Abstractive with Extractive Summaries}
\label{sec:method}
We introduce an intuitive and elegant pipeline that transforms an original abstractive gold-standard summary into its extractive version by selecting important sentences that \textit{match the expert's view of salience}, implicit in the human-written gold standard case summaries.
The proposed pipeline operates in two stages as described below.



\subsection{Candidate Sentence Selection}
During the first stage, we   construct  a set $C$  of the candidate sentences by identifying the high overlap sentences from the case document corresponding to each sentence in the abstractive gold standard summary. Let $D = \{s_1, s_2, \dots, s_n\}$ represent the sentences of the case document, and $G = \{a_1, a_2, \dots, a_m\}$ denote the sentence set of corresponding gold-standard abstractive summary (OAG). For each sentence pair $(s_i, a_j)$, we compute similarity between $s_i$ and $a_j$ using three ROUGE variants, R-1, R-2, and R-L. Subsequently, we compute relevance score \textbf{${\rho}_{ij}$} by averaging these three ROUGE scores. Thus, ${\rho}_{ij}$  captures the degree of lexical overlap between $s_i$ and $a_j$ at multiple levels of granularities. Subsequently, we identify top-\textit{k} scoring sentences from the case document corresponding to each $a_j (\in G)$, and added to $C$.  Using ROUGE metrics not only introduces objectivity in selecting the most informative sentences in the document, but also leads to a healthy selection of candidate sentences by ensuring that the selected judgment sentences align closely with opinion of salience in the OAG summary. Algorithm \ref{alg:sentence_selection} summarizes the sentence selection procedure.

\begin{algorithm}
\scriptsize
\caption{Candidate Sentence Selection}
\label{alg:sentence_selection}

\begin{algorithmic}[1]
\State \textbf{Input:} Sentence set of the case Document $D = \{s_1, s_2, \dots, s_n\}$,\\ Sentence set of the OAG  summary $G = \{a_1, a_2, \dots, a_m\}$, \\Selection threshold $k$
\State \textbf{Output:} Set of candidate sentences $C$
\State $C \gets \emptyset$ \Comment{Initialize empty candidate set}

\For{each summary sentence $a_j \in G$}
    \For{each document sentence $s_i \in D$}
        \State $r_1 \gets \text{R-1}(s_i, a_j)$; $r_2 \gets \text{R-2}(s_i, a_j)$; $r_L \gets \text{R-L}(s_i, a_j)$
        \State $\rho_{ij} \gets avg(r_1, r_2, r_L)$
    \EndFor
    \State $C \gets C \cup \text{Top-} k \text{ sentences from } D \text{ according to } \rho_{ij}$
\EndFor
\State \Return $C$
\end{algorithmic}
\end{algorithm}

\subsection{Summary Creation} 
In the second stage, we use the candidate set $C$ to generate the final extractive silver summary (TES). Since the set $C$ may contain some sentences with overlapping legal content, we use Maximal Marginal Relevance (MMR) method to iteratively select summary sentence from the candidate set $C$ \citep{MMR-1998}. The sentence is selected in a greedy manner that maximizes summary coverage while minimizing redundancy with the already selected sentences in the current summary. This technique is particularly advantageous in legal summarization, as it effectively distills a complex judgment into concise and non-repetitive summary while preserving both completeness and brevity in the final extractive version \citep{MMR-2011,2019-zhongAutomatic,billsum-2019,22-ExtandAbsMethods}.  Given the pool of candidate sentences $C$, the decision to include a sentence $c_{i} (\in C)$ is based on the sentence scoring using the following equation adapted from \citet{MMR-1998}.
\[
\text{MMR}(c_i) = \lambda \cdot \sigma(c_i,  C ) - (1 - \lambda) \cdot \sigma(c_i, S_{\text{E}})
\]

Here, $\sigma(c_i, C)$ measures the cosine similarity of candidate sentence $c_i$ to the set of sentences in $C$, and $\sigma(c_i, S_{\text{E}})$, measures the similarity between $c_i$ and the set of sentences already selected in the \textit{under-construction} extractive summary $(S_{\text{E}})$ to maintain diversity.  We set the parameter \( \lambda \) to 0.5 to balance relevance and diversity (the second term). The sentence with the highest marginal relevance score is \textit{greedily} chosen and added to the summary $S_{\text{E}}$ until its length approximately matches the word count of the OAG summary. 

\noindent
AugAbEx pipeline is an unsupervised,  transparent, scalable and explainable method, where the summaries are reproducible and inclusion of every sentence in TES can be explained. The pipeline can be extended to other domains for augmenting existing abstractive summarization datasets with extractive summaries.

%
\section{Multi-dimensional Evaluation Framework}
\label{sec:eval-metric}
\begin{figure*}[t]
    \centering
    \includegraphics[width=\textwidth]{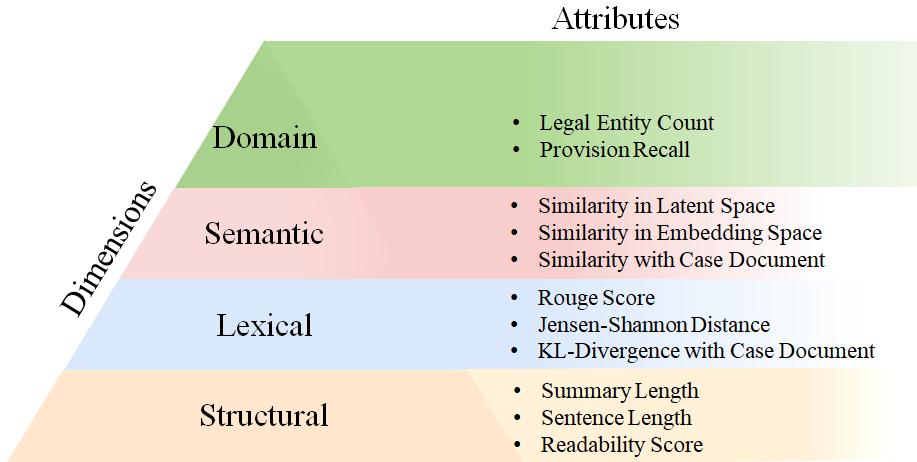} 
    \caption{Multi-dimensional automatic evaluation framework for transformed extractive silver (TES) summary}
    \label{fig:evaluationFramework}
\end{figure*}

For the transformed extractive summaries to be of practical use, the $\langle$OAG, TES$\rangle$  summary pair must be similar along the structural, lexical and semantic dimensions. Most importantly, a TES summary must preserve at least the same level of domain (legal) information as its corresponding OAG summary.  These requirements necessitate a multidimensional approach for the comparative assessment of the abstractive and the corresponding extractive silver case summaries. We compare OAG and TES summaries across domain, semantics, lexical, and structural dimensions, and assess multiple attributes for each dimension. We also employ Bradley-Terry test \citep{21-bt-test} to obtain robust statistical inference from the pairwise comparison of OAG and TES summary scores. Figure~\ref{fig:evaluationFramework} presents the blueprint of the evaluation framework, which is described in detail below.
\subsection{Domain Attributes}
\label{subsec:domain-attributes}
Human-written gold-standard summaries are dense in technical legal information because law experts skillfully distill essential legal terms, concepts, and entities entrenched in the judgment. This information is  crucial for  facilitating  correct and consistent interpretation of the case by law practitioners. 

Inclusion of sufficient and accurate domain-specific information in a judgment summary is the first and foremost trait of a high quality case summary. Legal entities are critically important for practitioners to assess contextual relevance and ensure alignment with other domain-specific information within the case document. Since a summary is inherently a lossy compression of the original document, the omission of legal terms, entities, and jargon contained in the source document must be minimal. Ergo, it is important to quantify the extent to which technical legal information is preserved during the transformation of OAG summaries. In order to assess the ability of the pipeline to preserve the core legal information, we  (i) count  all legal entities identified in the summary as the indicator of its richness for the end user, and (ii) compute recall  of legal provisions captured in TES summaries.  

The  legal entity count, \textit{Lent-cnt}  metric, is used to compare and quantify the difference in the core legal information contained in OAG and TES summaries.  A higher entity count in the extractive summary ensures that key legal information is secured, which provides clarity to a legal professional and confirms its fidelity. Applicable statutes and provisions mentioned in a case document include references to specific articles, sections, and clauses from laws, constitutions, and regulations. They  provide the legal framework to the court's decision and carry  vital information for interpretation and analysis of the case. We introduce \textit{ProvRecall} metric to measure the proportion of provisions present in the OAG summary that is preserved in the TES summary. If $P_O$ and $P_T$ denote the number of legal provisions identified in OAG and TES summary respectively, then the legal provisions recall is computed as $\textit{ProvRecall}  = \frac{P_O \cap P_T}{P_O}$. High recall indicates that the TES summary has secured a significant proportion of legal provisions. 
\subsection{Semantic  Attributes}
The next dimension of scrutiny delves into the quantitative assessment of how well the the transformed extractive silver (TES) summaries convey the core concepts, ideas, and information contained in the  original abstractive gold (OAG) summaries. We gauge the semantic similarity between the OAG and TES summaries using three-pronged approach. We compute the semantic similarity between the $\langle$OAG, TES$\rangle$ summary pair (i) within the latent space uncovered using latent semantic analysis (LSA) \citep{steinberger2009evaluation},  (ii) in the embedding space of the LegalBert \citep{legal-bert-2020}, and (iii) find the semantic similarity of two summaries with the case document using LSA. 
 
The semantic-based automatic similarity assessment metric using LSA \citep{steinberger2009evaluation} is reference-free and captures the commonality between the main topics of the two documents. Based on the distributional hypothesis, it effectively gauges the consistency and fidelity of the representation of a case document by the two summaries. We expand the analysis by computing the semantic alignment between the OAG and TES summaries in the embedding space using LegalBert \citep{legal-bert-2020}, which is an advanced embedding model trained to analyze the contextual relationships and semantic connections in legal documents. LegalBert embeddings capture the nuanced legal context and relationships within the text, which facilitates a sophisticated comparison of the semantic content of the two case summaries. Finally, we gauge the semantic congruence between the abstractive gold-standard reference summary and the transformed extractive silver summary by comparing them with the case document. We perform a paired comparison of the semantic proximity of the OAG and TES summaries to the case documents they represent to infer the performance of the AugAbEx pipeline. The three attributes offer a robust quantitative comparison of the information encased in the two summaries.
\subsection{Lexical Attributes} 
Considering  TES summaries as a practical proxy of OAG versions, the degree of lexical similarity between abstractive and extractive summaries indicates how well the key terms in the original abstractive summaries are preserved. A high overlap between $\langle$OAG, TES$\rangle$ summary pair reflects a consistent representation of the crucial content, enhancing the reliability of the transformed extractive silver summaries.

We use three lexical attributes to gauge the degree of shared textual matter between OAG and TES summaries and the difference between the usage of language. Specifically, we find the (i) vocabulary overlap between two summaries, (ii) distance between their term distributions, and (iii) comparative divergence of the two summaries from the case document. The classical ROUGE metric quantifies the vocabulary overlap between the two summaries. We employ ROUGE-\{1,2,L\} F-score with respect to OAG summaries, where R-1, R-2 capture informativeness, and R-L assesses sequence coherence. We take into account the probabilistic view of the terms in OAG and TES summaries, and use information-theoretic measures to compute distance and divergence between them. These measures provide insights into the writing style and emphasis, offering a perspective that extends beyond straightforward lexical overlap. Jensen-Shannon distance (JSD), a symmetric measure bounded between 0 and 1, quantifies the difference between term distributions of the two summaries. It is an interpretable, stable, and robust measure to compare the use of legal lexicon in OAG and TES summaries. The third lexical attribute, KL-divergence between the case document and the two summaries offers insights into language usage by revealing shared and distinct linguistic patterns.
\subsection{Structural Attributes}
Structural attributes offer a glimpse of the  surface-level characteristics of the textual data. The comparison between structural attributes of OAG and TES summaries furnishes insights into their distinct syntactical patterns. Considering \textit{terms} to be the minimal lexical units and sentences to be maximal lexical units, we use two basal metrics - (i) word count, which reflects the summary length,  (ii) average sentence length, which indicates the complexity of the sentences in the summary. We use a third metric, the Flesch-Kincaid reading ease score, which quantifies how easy the text is to read \citep{kincaid1975derivation}.  The reading score ranges from 0 to 100, with higher score indicating easier reading.  

Comparable summary length of OAG and TES summaries is integral for comparison at lexical, semantic and domain levels. The average sentence length indicates the summary's structural composition and granularity. With the objective to examine the impact of the transformation on the reading ease, we compare the readability of OAG and TES summaries using  Flesch-Kincaid reading ease score, which is computed as follows.
\begin{equation*}
FK\text{-}Score = 206.835 - 1.015 \times ASL - 84.6 \times ASW
\label{eq:fk-formula}
\end{equation*}
Here, \textit{ASL} is the average sentence length and \textit{ASW} denotes the average syllables per word. Note that the readability of the text with complex words is severely penalized due to the high negative coefficient of \textit{ASW}. Overall, texts with longer sentences and complicated  words tend to have lower readability scores, while texts featuring shorter sentences and less complex vocabulary obtain a higher readability score. 

\subsection{Statistical Test}
\label{subsec:statistical}
Aggregating the scores of the relevant metric using mean or median is the most popular strategy for comparison of two NLP systems. \citet{21-bt-test} argue that this procedure may be unreliable for two reasons. \textit{First}, the idiosyncrasies of these two aggregation function may obfuscate the true patterns, and \textit{second}, these aggregation functions ignore the instance-level pairing. The authors demonstrate, both theoretically and empirically, the robustness of the Bradley-Terry (BT) model, which analyses instance-level scores and aggregates to infer the relative superiority of one system over the other.

The BT model compares the evaluation scores of two systems at the instance level and infers the relative superiority of a system based on how frequently it wins. In this study, we compare systems $O$ and $T$, where $O$ represents the source of original abstractive gold (OAG) summaries, and system $T$ corresponds to the AugAbEx pipeline for generating extractive silver (TES) summaries.  Let $\lambda_O$ ($\lambda_T$) denote the number of times system $O$ ($T$) scores higher than system $T$ ($O$), over all instances in the dataset. Based on this data, BT model estimates the probability ($\hat\lambda_O$) as follows.
\begin{equation}
\label{eq:BT-model}
    P(O > T) = \frac{\lambda_O}{\lambda_O + \lambda_T}    
\end{equation}
The probability, $\hat\lambda_O$, connotes the estimated latent strength of the system $O$, and expresses the chance that it performs better than system $T$ for the dataset. The estimated strengths of the two systems are probabilities (Eq.~\ref{eq:BT-model}),  $\hat\lambda_O + \hat\lambda_T = 1$. Thus, a value $> 0.5$ is interpreted as higher strength.  We use the BT model\footnote{We use the publicly available implementation provided by the authors at \url{https://github.com/epfl-dlab/pairformance}.} to estimate the strengths of the systems $O$ and $T$ draw an inference about the leading system.

\subsection{Comparison with SOTA in Legal Case Summarization}
 We compare the quality of TES summaries against the summaries produced by recent baseline/state-of-the-art approaches in legal case summarization. We pick the best results reported  in recent research works and  compare the numbers against the performance of AugAbEx method.
\begin{enumerate}[label=\roman*)]
    \item \textit{Baseline methods}: LexRank \citep{2004-lexrank} and DSDR \citep{2012-dsdr-he} are established algorithms  for extractive summarization  that  have been examined for legal domain. These algorithms, evaluated for IN-Abs and UK-Abs datasets \citep{22-ExtandAbsMethods}, serve as strong baselines and are included as competitors due to their suitability for  use as extractive silver summary generators.
    \begin{enumerate}[label=\alph*)]
         \item \textit{LexRank} is a graph based method that computes sentence importance using eigenvector centrality. It constructs a sentence similarity graph where cosine similarity between sentences enable highly connected sentences to receive higher importance scores. 
         \item DSDR  models the relationships among sentences using linear and nonnegative linear reconstruction objectives, and generates summary by selecting sentences that can best reconstruct the original document. 
    \end{enumerate}
    \item \textit{State-of-the-art approaches}: The following three SOTA methods closely align with our work, as they leverage the original abstractive gold summary (OAG) as the first step towards high quality summarization. The three methods are evaluated on BillSum dataset. 
     \begin{enumerate}[label=\alph*)]
          \item \textit{Athena} \citep{2023-hybrid-align-moro} is a joint segmentation-summarizer approach, which learns informative text representation using an \textit{align-then-abstract} strategy. The approach aligns document chunks with OAG sentences by maximizing semantic similarity between $\langle$\textit{chunk, sentence}$\rangle$ pairs, while simultaneously optimizing segmentation quality through alignment loss. Subsequently, BART-based abstractive summarizer uses aligned representations for summary generation.
        \item \textit{DCESumm} \citep{2024-hybrid-DCESumm} is a two step summarization approach, which  combines supervised sentence-level relevance prediction and unsupervised clustering mechanism. The  method first identifies summary-worthy sentences by measuring the similarity between abstractive summary (OAG) sentences and document sentences using contextual representation from a pretrained language model. Subsequently, it clusters similar sentences and enhance their scores to create an extractive summary. 
        \item \textit{SENDE} \citep{2026-soft-labeling-sende} is a three step hybrid extractive summarization method that first identifies salient sentences from the document using ROUGE- relevance scoring with respect to the OAG sentences. Subsequently, the sentence representation is improved through TSDAE-based unsupervised fine-tuning of BERT/RoBERTa models, and finally an extractive summary is generated using a CNN-based model.
    \end{enumerate}
\end{enumerate}
\section{Corpora}
\label{sec:datasets}
This section briefly describes seven English case summarization datasets that are augmented with extractive silver summaries. Drawn from the judgment-summary corpora from four jurisdictions — India, U.S.,  U.K., and Australia, these datasets offer diverse legal documents across different legal systems. Below is a brief description of each dataset. 
\begin{enumerate}[label=\roman*)]
     \item IN-Jud-Cit is a small dataset comprising judgments from Indian courts related to IPR. The dataset was originally curated for citation-based summarization \citep{purnima-etal-2023-citation}.
     \item ILC dataset consists of a collection of judgments and their corresponding  abstractive summaries. These summaries are curated from various sources such as Briefcased, Primelegal, Indian Kanoon, Lawtimes Journal, and respective High Court websites \citep{ILC}.
    \item IN-Abs dataset consists of  case documents from the Supreme Court of India and their corresponding abstractive summaries from the Legal Information Institute of India \citep{22-ExtandAbsMethods}.
    \item CivilSum, the largest legal corpus available in India consists of judgments from Supreme Court of India. A distinctive characteristic of this dataset is that the gold standard abstractive summaries are formatted as phrases, rich in legal terms and entities rather than complete sentences. The summaries include references to the original judgment paragraphs containing the phrases \citep{2024-civilsum}.
    \item UK-Abs dataset features comparatively lengthier judgments, and their corresponding official press summaries from the UK Supreme Court. The summaries are abstractive and structured into three segments: `Background to the Appeal', `Judgement', and `Reasons for Judgement' \citep{22-ExtandAbsMethods}.
    \item Australian dataset is one of the earliest dataset available from the Federal Court of Australia to summarize legal documents. It is characterized by notably shorter judgments and summaries which are formatted as phrases \citep{Australian-dataset}.
    \item BillSum dataset is a comprehensive collection of US Congressional and California state bills, accompanied by human-written summaries as phrases. The documents are well structured and are organized into sections with titles, composed of phrases or sentences \citep{billsum-2019}. 
\end{enumerate}
 \begin{table*}[t]
    \centering
    \caption{Statistical properties of  the datasets. WC: Average word count (summary length), SC: Average sentence count in the document, D: Case documents, S: Summary, CR: Average compression ratio of the dataset}  
    \label{tab:stats-datasets}
    \begin{tabular}{@{}lrrrrrr@{}}
        \toprule
         &   &   \multicolumn{2}{c}{\textbf{WC}} &  \multicolumn{2}{c}{\textbf{SC}} &     \\ \cmidrule{3-4} \cmidrule{5-6} 
        \textbf{Dataset} &   \textbf{\#D} & \textbf{D}  & \textbf{S} & \textbf{D} & \textbf{S} & \textbf{CR} \\ \midrule
        \textbf{IN-Jud-Cit}  & 50             & 21975 & 528  & 643  & 24  &  45.53
\\
        \textbf{ILC}         & 3073           & 2339  & 561  & 74   & 17  & 4.72
\\
        \textbf{In-Abs}      & 7128           & 4376  & 842  & 138  & 29 & 6.02
\\
        \textbf{CivilSum}    & 23095          & 2110  & 104  & 83   & 5   & 21.7
\\
        \textbf{UK-Abs}      & 793            & 14267 & 1098 & 442  & 41   & 12.9
\\
        \textbf{Australian}  & 3890           & 6388  & 65    & 344  & 8   &  125.91
\\
        \textbf{BillSum}     & 22218          & 1259  & 176  & 30   & 5  & 10.74
\\ 
        \bottomrule
    \end{tabular}%
    \end{table*}   
 Table \ref{tab:stats-datasets} summarizes the key statistics of the judgments and their corresponding abstractive summary for each dataset.  With wide-ranging document lengths and compression ratios\footnote{The compression ratio  for each document is defined as the ratio of the document length to the  gold summary length (in words). The average compression ratio (CR) is computed by averaging the CR values of all documents in the dataset.}, the diversity among documents in the datasets strengthens the comprehensive comparative analysis and evaluation of transformed extractive summaries. 
\section{Experimental Results and Discussion}
\label{sec:expt}
We report the results of the comprehensive assessment of the TES summaries against the OAG summaries for four facets (Sections~\ref{sec:domain-level-analysis} - \ref{sec:structural-level-analysis}) using the evaluation framework described in Section \ref{sec:eval-metric}. Furthermore, we present a comparison of the AugAbEx pipeline with strong baselines and state-of-the-art approaches in Section~\ref{sec:comparison-sota}.
\begin{figure}[h]
    \centering
    \includegraphics[width=0.7\linewidth]{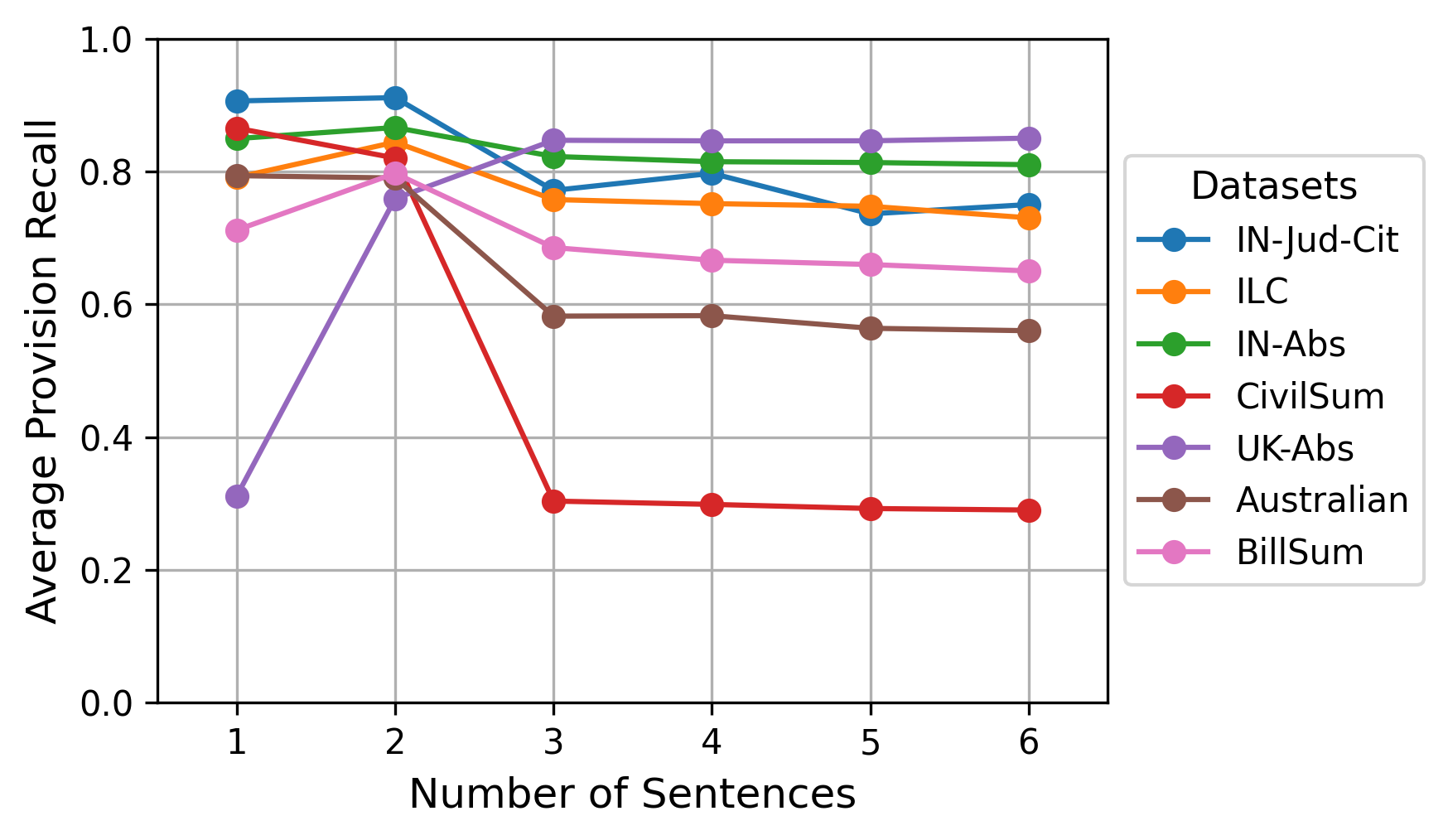} 
    \caption{ Macro-averaged recall score of provisions in the transformed extractive silver summaries for varying number of candidate sentences ($k$).}
    \label{fig:provRecall_scores_plot}
\end{figure}
We empirically determine the  value of $k$ for selecting candidates  corresponding to each OAG summary sentence (Step 11 in Algorithm \ref{alg:sentence_selection}), by varying the value of $k$ from $1 - 5$ (Fig. \ref{fig:provRecall_scores_plot}).  We consider $ProvRecall$ as a measure to be maximized due to the importance of technical legal information  for legal experts. We observe that $k=2$  yields the best  score for most datasets and plateaus thereafter  and the candidate pools are created accordingly.
\subsection{Domain-Specific Analysis}
\label{sec:domain-level-analysis}
   To extract legal entities from OAG and TES summaries, we leverage the LegalNER\footnote{Model downloaded from \url{https://github.com/Legal-NLP-EkStep/legal_NER}} model, trained by \citet{2022-legal-ner} for Indian legal documents.   The extracted \textit{entities} and  the \textit{legal provisions} for OAG and TES summaries  are analyzed below as the metric introduced in Section \ref{subsec:domain-attributes}.
 \begin{table}[h]
    \centering
    \caption{Mean and median counts (\textit{Lent{-}cnt}) of Legal-NER entities in OAG and TES summaries for all datasets. ${\hat{\lambda}_T}$: Estimated strength of system \textit{T}. }
    \label{tab:bt-score-Legal-Ner-entity}
    \begin{tabular}{lcccccc} 
    \toprule
     & \multicolumn{2}{c}{Mean}& \multicolumn{2}{c}{Median}&\multicolumn{1}{c}{} & \multicolumn{1}{c}{}\\ 
    \cmidrule{2-3} \cmidrule(l){4-5} 
    Dataset & OAG & TES & OAG & TES & $\hat{\lambda}_T$ & ProvRecall\\ 
    \midrule
    IN-Jud-Cit & 8.05  & 9.06  & 6.56  & 7.75  &  0.6539 & 0.9110\\
    ILC        & 12.11 & 11.53 & 11.00 & 10.73 &  0.4317  & 0.8440\\
    IN-Abs     & 15.43 & 16.77 & 11.99 & 13.99 &  0.6423 & 0.8660\\
    CivilSum   & 3.12  & 2.75  & 3.00  & 2.00  &  0.3557 & 0.8200\\
    UK-Abs     & 19.59 & 20.55 & 18.84 & 19.61 &  0.5647 & 0.7590\\
    Australian & 0.67  & 2.26  & 0     & 2.00  &  0.8859  & 0.7900\\
    BillSum   & 3.41   & 3.78  & 3.00  & 3.00  &  0.5210 &0.7970\\ 
    \bottomrule
    \end{tabular}
 \end{table}
\begin{enumerate}[label=\roman*)]
    \item \textit{Legal Entity Count}: Table \ref{tab:bt-score-Legal-Ner-entity} shows the \textit{mean} and \textit{median} counts of extracted legal entities (\textit{Lent-cnt}) from OAG and  TES summaries. Both counts for TES summaries are generally higher than  OAG summaries for all datasets except  ILC and CivilSum. This is unexpected, especially for ILC dataset, which has conventional  human-written summaries unlike phrasal summaries of CivilSum dataset. In-depth analysis of ILC summaries failed to reveal a convincing explanation for the low \textit{Lent-cnt} values. For the CivilSum dataset, the phrases abstracted by the law experts carry  more legal information and a higher concentration of legal entities in OAG summaries, boosting  \textit{Lent-cnt} of OAG summaries.
    
    Interestingly, for non-Indian case documents,   more   legal entities are identified in TES than in OAG summaries.  The mean and median counts for  Australian and BillSum dataset are  lower than   those for UK dataset.   Apparently, the LegalNER model trained using Indian judgments is not able to identify the entities prevalent in the Australian and BillSum documents, but does well for UK documents. A possible explanation is the historical connection between the Indian and UK judicial systems, due to which the usage of legal language and substantial terminology is common in the two systems.
    
    Since the difference in the mean and median scores of the two summaries is \textit{small} for all datasets, it is prudent to perform pairwise comparison for individual case documents to statistically infer the relative strengths of the two systems, computed using Eq.~\ref{eq:BT-model}. The second last column of Table \ref{tab:bt-score-Legal-Ner-entity} shows the estimated latent strength of system \textit{T} computed by BT model (Section \ref{subsec:statistical}). 
    It is evident that system \textit{T} exhibits higher  strength in  agreement with the mean and median counts for five datasets, and lower strength for ILC and CivilSum dataset in consonance with the corresponding lower mean and median values. Thus, there is a higher probability of system \textit{T} summaries carry more legal entities for five datasets.      
    \item \textit{Provision Recall}: The last column of Table \ref{tab:bt-score-Legal-Ner-entity} shows the \textit{ProvRecall} metric for all datasets. It is evident that the provision recall for the four Indian datasets is high as compared to non-Indian datasets. This indicates that TES summaries are nearly as rich in legal provisions as OAG summaries for Indian datasets, and a significant amount of important legal information is preserved during transformation.  
\end{enumerate}
\textit{ The results reveal the richness of TES summaries in core legal information, as the AugAbEx pipeline prioritize sentences rich in  legal information in consonance with the experts' knowledge preserved in the OAG summaries.}
\subsection{Semantic Analysis} 
\label{sec:semantic-level-analysis}
Semantic comparison of the OAG and TES summaries aims to understand how different or similar are the meanings conveyed by the two texts. Table \ref{tab:semanticSimilarity-OAG-TES} shows the macro-averaged semantic similarity scores between OAG and TES summaries across various datasets, in the latent and embedding spaces.

\begin{enumerate}[label=\roman*)] 
        \begin{table*}[]
        \caption{Macro-averaged semantic similarity between OAG and TES summaries in latent and embedding space along with standard deviation. SS: Semantic Similarity, L-S: Latent space revealed by LSA, E-S: Embedding space of LegalBert}
        \label{tab:semanticSimilarity-OAG-TES}
        \centering
        \resizebox{\textwidth}{!}{%
        \begin{tabular}{@{}lccccccc@{}}
        \toprule
        \textbf{SS}  &
          \textbf{IN-Jud-Cit} &
          \textbf{ILC} &
          \textbf{IN-Abs} &
          \textbf{CivilSum} &
          \textbf{UK-Abs} &
          \textbf{Australian} &
          \textbf{BillSum} \\ \midrule
        \textbf{L-S}&
           0.9238 ± 0.06 &
          0.8855 ± 0.10 &
          0.9374 ± 0.05 &
          0.7164 ± 0.15 &
          0.9319 ± 0.03 &
          0.6306 ± 0.20 &
          0.8240 ± 0.13 \\  \midrule
        \textbf{E-S}&
           {0.9801 ± 0.01} &
           {0.9774 ± 0.01} &
           {0.9856 ± 0.01} &
           {0.9537 ± 0.02} &
           {0.9834 ± 0.01} &
           {0.9062 ± 0.05} &
          {0.9582 ± 0.03} \\
          \bottomrule
        \end{tabular}%
        }
        \end{table*}
        \item \textit{Semantic Similarity in Latent Space:} The first row of the Table \ref{tab:semanticSimilarity-OAG-TES} shows semantic similarity  of the OAG and TES summaries in latent space using LSA method \citep{steinberger2009evaluation}. The results demonstrate consistently high similarity scores for all datasets. However, CivilSum, Australian, and BillSum datasets achieve comparatively lower scores, which is in-line with the lower \textit{Lent-cnt} reported in Table~\ref{tab:bt-score-Legal-Ner-entity}. This indicates that TES summaries effectively capture key topics and legal information similar to those in OAG summaries for five datasets, but moderately for  CivilSum and Australian datasets.
%
        \begin{figure}[h]
            \centering
            \includegraphics[width=1\linewidth]{figures/violin-similarityScore-judgment-OAG-TES.png}
            \caption{Distributions of semantic similarity scores of OAG and TES summaries with case documents in latent space.}
            \label{fig:semantic-similarity-judgment-latentspace-OAG-TES}
        \end{figure}
        \item \textit{Semantic Similarity in Embedding Space:}  Second row of Table \ref{tab:semanticSimilarity-OAG-TES} shows the similarity scores of OAG summaries and TES summaries using the LegalBert embeddings. The scores assert high semantic similarity across all datasets, suggesting that TES summaries closely mirror the legal information contained in the OAG summaries.  Notably, the scores are uniformly higher than those in latent space, implying that LegalBert provides a more precise measurement of semantic alignment in legal contexts\footnote{Appendix \ref{a:contrast-latent-embed} shows a comparison  between the distribution of similarity scores of the summaries in the latent and embedding space of LegalBert.}. The strong semantic alignment underscores the robustness of the proposed summarization pipeline in preserving the legal content of the original case documents. Interestingly, there is high semantic similarity between OAG and TES summaries for CivilSum, Australian and BillSum datasets, comparatively low legal information. High scores for the CivilSum and Australian datasets in the legal embedding space suggests that despite the phrasal form of OAG summaries, the pipeline successfully draws a decent quantum of technical legal information in the form of sentences.  
%
        \begin{table*}[h]
        \caption{Macro-averaged semantic similarity Score of OAG and TES summaries relative to case documents using LSA.  CD-OAG: Case Document Vs. OAG, CD-TES: Case Document Vs. TES, $\hat \lambda_T$: Estimated strength of system \textit{T}.}
        \label{tab:OAG-TES-similarity-score-judgment}
        \centering
        \resizebox{\textwidth}{!}{%
        \begin{tabular}{@{}lccccccc@{}}
        \toprule
        \textbf{Dataset}& \textbf{IN-Jud-Cit}& \textbf{ILC}& \textbf{IN-Abs}& \textbf{CivilSum} & \textbf{UK-Abs} & \textbf{Australian} & \textbf{BillSum} \\ \midrule
        
        \textbf{CD-OAG} & 0.8464  & 0.8454  & \textbf{0.8993  } & 0.6599  & 0.8908  & 0.5178  & 0.7724   \\
        
       \textbf{CD-TES } & \textbf{0.8541  }& \textbf{0.8817  } & 0.8918  & \textbf{0.6836 } & \textbf{0.8955  } &  \textbf{0.5671  } & \textbf{0.8208  } \\

        {$\bm{\hat{\lambda}_T}$}& 0.6208& 0.6506& 0.3120& 0.5732& 0.5425& 0.6093& 0.6689\\ \bottomrule
        \end{tabular}%
        }
        \end{table*}     
        \item \textit{Semantic Similarity with case documents:} Figure \ref{fig:semantic-similarity-judgment-latentspace-OAG-TES} shows the comparative distributions of semantic similarity scores between case documents and their corresponding OAG and TES summaries, in the LSA latent space. The distributions are largely similar with the quartiles of both distributions placed reasonably high for all datasets. These observations suggest that TES summaries preserve the semantics of the case documents faithfully. Table~\ref{tab:OAG-TES-similarity-score-judgment} shows the macro-average semantic similarity scores of OAG and TES summaries relative to the case documents. The TES summaries achieve higher scores for all datasets except for the IN-Abs dataset, where the scores are comparable. We perform the BT test, using Eq.~\ref{eq:BT-model}, to  gain further insights into performance of the systems $O$ and $T$.
        
         Figure~\ref{fig:SS-Abs-Billsum} presents the contour plots of the joint density distribution of  paired scores for semantic similarity  between the summaries and case documents for IN-Abs and BillSum datasets. The two datasets are chosen to contrast the strengths of system $T$. In Fig. \ref{fig:IN-Abs-joint-density-plot-SS} (IN-Abs dataset), a large portion of mass clustered in lower triangle (shaded region) indicates that there are more instances for which system $O$ scores better. However, the inset table shows that there is a marginal difference in mean and median of both systems. On the other hand, Fig. \ref{fig:BillSum-joint-density-plot-SS} exhibits  more mass concentrated in upper triangle  for BillSum dataset, indicating that system $T$ scores are better (for $\approx 67\%$ instances as shown in Table~\ref{tab:OAG-TES-similarity-score-judgment}). The inset table shows that the mean and median are higher for system $T$ by a reasonable gap. Consequently, it is reasonable to conclude that TES summaries depict comparable content similarity to the complete case document as the original abstractive summaries for all datasets. 
          \begin{figure*}[t]
            \centering
            \begin{subfigure}[b]{0.45\linewidth}
                \centering
                \includegraphics[width=\linewidth]{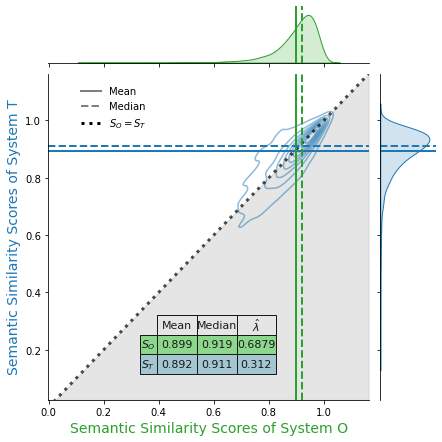}
                \caption{IN-Abs}
                \label{fig:IN-Abs-joint-density-plot-SS}
            \end{subfigure}
            \hfill
            \begin{subfigure}[b]{0.45\linewidth}
                \centering
                \includegraphics[width=\linewidth]{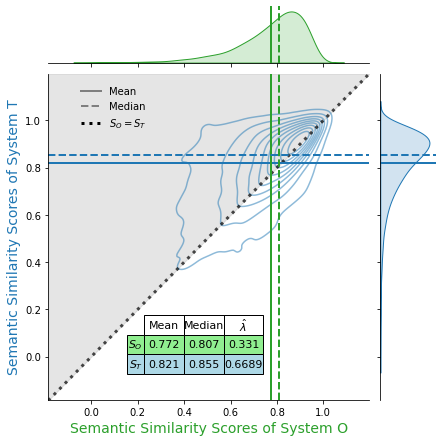}
                \caption{BillSum}
               \label{fig:BillSum-joint-density-plot-SS}
            \end{subfigure}
            \caption{Joint density plots of the Semantic Similarity scores of OAG and TES summaries with judgment for (a) IN-Abs and (b) BillSum datasets. $S_O$: System \textit{O}, $S_T:$ System \textit{T} }
            \label{fig:SS-Abs-Billsum}
        \end{figure*}    
\end{enumerate}
\textit{The results establish that the TES and OAG summaries are equally informative in the semantic dimension, and the transformed extractive silver summaries are of sufficient quality for training high-quality hybrid and extractive case summarization models. }  
\subsection{Lexical Analysis} 
\label{sec:lexical-level-analysis}
The comparison of OAG and TES summaries for the lexical attributes is reported in Tables~\ref{tab:Rouge-JSD-OAG-TES} and~\ref{tab:KL-divergence} . ROUGE metric quantifies the lexical overlap, while Jensen Shannon distance (JSD) quantifies the differences between term distributions in OAG and TES summaries.  The comparison of KL-divergence of the OAG and TES summaries from the case documents presents insights into the subtle differences at the lexical level. We discuss below the results for the three attributes.
\begin{table}[t]
\centering
\caption{Comparison of (i) lexical overlap  of OAG and TES summaries across all datasets using macro-averaged ROUGE F-measure scores, (ii) JSD: Jensen Shannon Distance between OAG and TES summaries. }
\label{tab:Rouge-JSD-OAG-TES}
\begin{tabular}{@{}lccccccc@{}}
\toprule
\textbf{Dataset} & \textbf{IN-Jud-Cit} & \textbf{ILC}  & \textbf{IN-Abs} & \textbf{CivilSum} & \textbf{UK-Abs} & \textbf{Australian} & \textbf{BillSum}  \\
\midrule
\textbf{R-1} & 70.30 & 68.01  & 71.37  & 45.21    & 67.67    & 38.10  & 55.19    \\
\textbf{R-2} & 47.14 & 45.22 & 46.91  & 18.78    & 36.89    & 15.55   & 31.77     \\
\textbf{R-L} & 73.03 & 69.72 & 72.83   & 42.75   & 69.67     & 41.63   & 54.08  \\
\midrule
\textbf{JSD} & 0.1806 & 0.2129  & 0.1631   & 0.3802   & 0.1718   & 0.4366   & 0.3008    \\
\bottomrule
\end{tabular}%
\end{table}
\begin{enumerate}[label=\roman*)]
    \item \textit{Lexical Overlap:}  First three rows of Table \ref{tab:Rouge-JSD-OAG-TES} show the lexical overlap between the  original abstractive gold standard and transformed extractive summaries in terms of unigrams (R-1), bigrams (R-2) and longest sub-sequence (R-L). Reasonably high scores for five datasets indicate that the transformed extractive summaries effectively capture the key content of the abstractive summaries and retain essential term-level information. The Australian and CivilSum datasets stand out with relatively lower ROUGE scores. The discrepancy is attributed to the nature of the summaries therein, which are predominantly phrasal and have reduced word overlap compared to sentential summaries for other datasets.
    \item \textit{Difference in Term Distribution:} Jensen-Shannon distance (JSD) between the TES and OAG summaries (bottom row of Table \ref{tab:Rouge-JSD-OAG-TES}) quantifies the difference between the term distribution in summaries. Smaller value of JSD affirms that the content is well preserved during transformation. When cross-referenced with the ROUGE scores, the JSD provides an additional layer of comparative assessment.
   
    Five datasets (IN-Jud-Cit,  ILC, IN-Abs, and UK-Abs) exhibit reasonably close alignment of the term distributions in the OAG and TES summaries, suggesting that the transformed extractive summaries (TES) effectively preserve the content of the original abstractive summaries (OAG). However, CivilSum, Australian, and BillSum datasets show relatively higher JSD scores, asserting major differences in the vocabularies. The observation reinforces the low lexical overlap for the two datasets revealed by the ROUGE metric, which is attributable to the unconventional writing style  of the original abstractive summaries.
     \begin{figure}[t]
            \centering
            \includegraphics[width=0.8\linewidth]{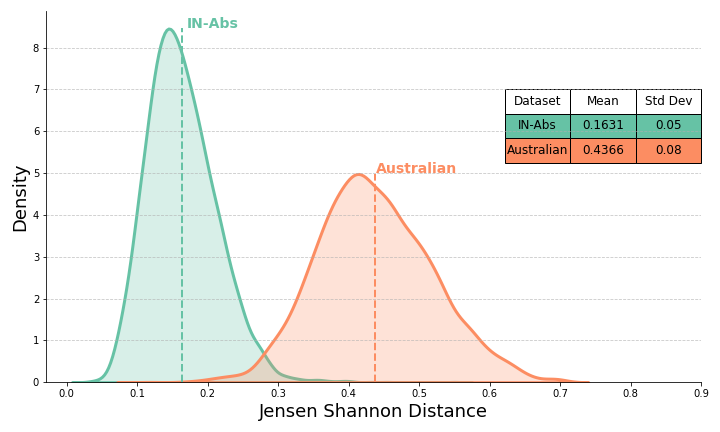} 
            \caption{Density plot of JS distances between the OAG and TES summaries for IN-Abs and Australian datasets. The dotted lines show the mean of the distances.}
            \label{fig:jsd-terms-2-datasets-density-plot}
    \end{figure}
    \begin{table*}[t]
    \caption{Macro-averaged KL-divergence scores of OAG and TES summaries for the case documents for all datasets, $\hat \lambda_T$: Estimated strength of system \textit{T}. Low divergence (better) scores are highlighted in bold.}
    \label{tab:KL-divergence}
    \centering
    \begin{tabular}{lccc}
    \toprule
    \textbf{Dataset} & \textbf{OAG} & \textbf{TES} & \textbf{$\bm{\hat{\lambda}_T}$}\\ 
    \midrule  
    \textbf{IN-Jud-Cit} & 1.88 ± 0.53          & \textbf{1.69 ± 0.52} & 0.8846 \\
    \textbf{ILC}        & 0.91 ± 0.44          & \textbf{0.89 ± 0.46} & 0.6737 \\
    \textbf{IN-Abs}     & 0.99 ± 0.41          & \textbf{0.94 ± 0.42} & 0.5634 \\
    \textbf{CivilSum}   & \textbf{1.78 ± 0.58} & 2.14 ± 0.63          & 0.2018 \\
    \textbf{UK-Abs}     & 1.22 ± 0.28          & \textbf{1.12 ± 0.26} & 0.8566 \\
    \textbf{Australian} & \textbf{3.08 ± 0.86} & 3.28 ± 0.87          & 0.4042 \\ 
    \textbf{BillSum}    & 1.43 ± 0.55 & \textbf{1.32 ± 0.62} & 0.6856 \\ 
    \bottomrule
    \end{tabular}%
\end{table*}

    We  plot the density curve of the JSD scores for IN-Abs and Australian datasets in Fig. \ref{fig:jsd-terms-2-datasets-density-plot}, to contrast their distributions. Note that the two datasets lie at the extreme ends of the spectrum of JSD score distributions. The density curve for TES summaries for the IN-Abs dataset exhibits the least JS distance and variance, while the JSD scores of the Australian dataset are larger and vary more. Thus the TES versions for Australian dataset summaries bear lower vocabulary overlap with their OAG counterparts. Both curves are slightly right skewed, with the mean (vertical line) falling on the right of the peak. Density curves of JSD scores for other datasets are shown in Appendix \ref{a:jsd-all-density}.

    \item \textit{Comparative Divergence from the case documents:} We find the KL-Divergence (KLD) of both summaries from the case documents and analyze the paired performance of the two systems at the instance level using BT model (Section \ref{subsec:statistical}). Table~\ref{tab:KL-divergence} shows that the  mean divergence of TES summaries is smaller for all, barring  CivilSum and Australian datasets. Note that divergence is a negative trait, and a lower score for TES summaries is favorable. The estimated strength of system $T$, shown in the table is higher for all, except the two datasets. This implies that the  TES summaries for most  datasets are lexically closer to the case documents.  The Australian and CivilSum  datasets, the OAG summaries of which have similar pathology, exhibit similar patterns of higher strength for system $O$. 
\begin{figure*}[th]
    \centering
    \begin{subfigure}[b]{0.45\linewidth}
        \centering
        \includegraphics[width=\linewidth]{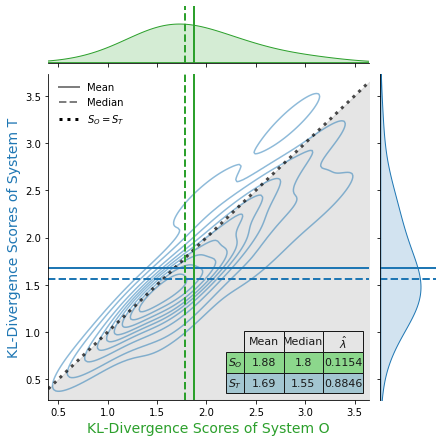}
        \caption{IN-Jud-Cit}
        \label{fig:IN-Jud-Cit-joint-density-plot}
    \end{subfigure}
    \hfill
    \begin{subfigure}[b]{0.45\linewidth}
        \centering
        \includegraphics[width=\linewidth]{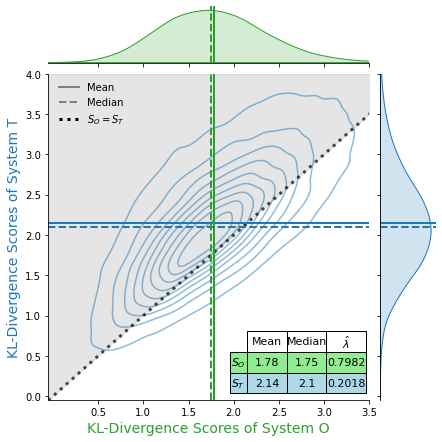}
        \caption{CivilSum}
       \label{fig:civilsum-joint-density-plot}
    \end{subfigure}
    \caption{Joint density plots of the KL-Divergence scores of OAG and TES summaries for (a) IN-Jud-Cit and (b) CivilSum datasets. $S_O$: System \textit{O}, $S_T:$ System \textit{T}}
    \label{fig:kld-cit-civilsum}
\end{figure*}
Figure \ref{fig:kld-cit-civilsum} shows the contour plots of the joint density of KLD scores of the two systems for IN-Jud-Cit and CivilSum datasets, which exhibit extreme latent strengths of system \textit{T} in this experiment.  More mass is concentrated in the lower  region for the IN-Jud-Cit dataset, which means that there are more instances for which the divergence of OAG summaries from the case documents is higher. System $T$ scores lower (better) for approximately  $88\%$ instances as suggested by $\hat\lambda_T$ in Table \ref{tab:KL-divergence}. The inset table shows that both mean and median divergence are lower for system $T$ for this dataset. On the other hand, more mass is concentrated in the upper  region for CivilSum summaries, indicating that the extractive summaries of system $T$ demonstrate higher divergence with respect to case documents and differ more from the original summary at lexical level. The mean and median KLD scores are lower for system $O$, indicating higher strength for CivilSum dataset.
\end{enumerate}

\noindent
\textit{Analysis of the results of three lexical attributes affirms strong lexical similarity between OAG and TES summaries for most datasets.  The structural differences in CivilSum and Australian datasets are reflected in their lexical variations, as expected.} 

\subsection{Structural Level Analysis}
\label{sec:structural-level-analysis}
Table \ref{tab:structural-attributes} shows the comparison of the basic structural attributes of the original abstractive gold  (OAG) and transformed extractive silver (TES) summaries. Scores for OAG summaries are placed outside the parenthesis, and those for TES are in.  We observe the following from the table.
\begin{table*}[htb]
\caption{Comparison of basic structural attributes and readability scores of OAG and TES summaries. Numbers inside parenthesis are for TES summaries, and those outside are for OAG. WC: word count (summary length) in words, Sent-Len: sentence length in words,  FK-Score: Flesch Kincaid readability score.}
\label{tab:structural-attributes}
\centering
\resizebox{\textwidth}{!}{%
\begin{tabular}{@{}lccccccc@{}}
    \toprule
    \textbf{} &
      \textbf{IN-Jud-Cit} &
      \textbf{ILC} &
      \textbf{In-Abs} &
      \textbf{CivilSum} &
      \textbf{UK-Abs} &
      \textbf{Australian} &
      \textbf{ BillSum} \\ \midrule
    \textbf{Avg. WC }  & 528 (533)     & 561 (551)   & 842 (861)  & 104 (119)   & 1098 (1114)      & 65 (79)  & 176 (217)   \\
    \textbf{Med. WC }   & 465 (479)     & 485 (505)  & 638 (661)   & 92 (107)    & 1086 (1103)  & 48 (62)   & 155 (192)      \\ \midrule
    \textbf{Avg. Sent-Len } & 23 (29)   & 34 (35)   & 29 (32)          & 25 (30)     & 28 (29)      & 9 (22)   & 38 (55)         \\
    \textbf{Med. Sent-Len }  & 21 (25)   & 28 (28) & 24 (27)     & 19 (25)     & 24 (25)           & 6 (19)    & 31 (43)   \\ \midrule
    \textbf{FK-Score} &
      49.9 (48.33) &
      48.83 (48.32) &
      51.21 (51.86) &
      27.61 (44.36) &
      47.98 (48.11) &
      -10.57 (47.13)  &
      10.19 (12.79)\\ \bottomrule
    \end{tabular}%
    }
\end{table*}
\begin{enumerate}[label=\roman*)]
    \item \textit{Summary length:} Average and median lengths (in words) of OAG and TES summaries are reported in the first two rows of Table \ref{tab:structural-attributes}. The values indicate that the transformed extractive summaries are of nearly same length compared to the abstractive gold summaries for IN-Jud-Cit, ILC, IN-Abs and UK-Abs datasets. For the remaining three datasets (CivilSum, Australian and BillSum), the transformed summaries are nearly 20\% longer. The underlying reason for shorter  original abstractive summaries is their phrasal structure. The corresponding extractive summaries tend to be longer, as they are composed of complete sentences. 
    \item \textit{Sentence length:} We report the average and median sentence lengths for the OAG and TES summaries (Rows 3 and 4) and observe that these values for TES summaries are marginally higher than those of OAG summaries. Since, the latter are written by human experts, the sentences are well condensed and are therefore shorter. CivilSum, BillSum  and Australian datasets exhibit longer extractive summaries, which is again due to the OAG summaries being structured as phrases, unlike the TES summaries, which are complete sentences. 
    \item \textit{Reading ease:} The bottom row in Table \ref{tab:structural-attributes} shows macro-averaged Flesch-Kincaid readability  scores of OAG and TES summaries. Both type of summaries have comparable readability levels, except for the CivilSum and Australian datasets. The disparity in the macro-averaged reading score of  CivilSum abstractive summaries (FK score - 27.61) versus the extractive summaries (FK score - 44.36) is attributed to the phrasal structure of the OAG summaries. Hence, they are harder to read compared to the TES summaries. 
    
    Interestingly, the Australian dataset exhibits a negative score, which is due to few sentences in OAG summaries. A negative Flesch-Kincaid readability  score is typically found in technical, academic, or legal documents. Such documents are challenging to read and require advanced reading skills.  Note that TES summaries for the same dataset, with FK score of 47.13, are more reader-friendly and highlight a more accessible writing style.
\end{enumerate}
\textit{We conclude that the three structural attributes for OAG and TES summaries in five datasets are comparable. However, CivilSum and Australian datasets stand out due to the unconventional phrasal structural form of the original abstractive summaries, which are crafted by human experts. As seen earlier, these datasets may exhibit different trends for lexical, semantic and domain-specific attributes. One example summary from each of the CivilSum and Australian datasets is shown in Appendix \ref{a:example-phrasal}.}
\subsection{Comparative Analysis with SOTA Legal Summarization Methods}
\label{sec:comparison-sota}
We present the comparative performance of AugAbEx  pipeline for IN-Abs, UK-Abs, and BillSum (US) datasets, which are adopted in recent state-of-the-art (SOTA) legal summarization studies. These approaches report ROUGE metric on test datasets of varying sizes. Table~\ref{tab:comparison-existing-methods} shows the ROUGE-\{1,2,L\} F-scores reported by the corresponding papers for the three datasets. 
\begin{table*}
\caption{Comparative performance of AugAbEx pipeline against baseline and state-of-the-art approaches. The SOTA results are reported from the referenced papers and best scores are in bold. $^\dagger$: evaluated on test set of 100 documents, $^{*}$: evaluated on full corpus, S: Supervised, U: Unsupervised, Ex: Extractive, Ab: Abstractive.}
\label{tab:comparison-existing-methods}
\begin{subtable}{0.7\textwidth}
\subcaption{Comparison of AugAbEx method with baseline methods}
\label{tab:a-comparison-with-baseline}
\begin{tabular}{llcccccc} \toprule
           & Approach &  \multicolumn{3}{c}{IN-Abs}&  \multicolumn{3}{c}{UK-Abs}\\\midrule
           && R-1&  R-2 &  R-L&  R-1&  R-2&  R-L\\ \midrule
             Baseline & LexRank$^\dagger$ (U, Ex)& 43.60& 19.50& 28.40 &  48.10& 18.70 & 26.50\\
  &\citep{22-ExtandAbsMethods}& & & & & & \\
          &DSDR$^\dagger$ (U, Ex)& 48.50& 22.20& 27.00&  48.40 & 17.40& 22.10\\ 
  &\citep{22-ExtandAbsMethods}& & & & & & \\ [0.13cm]
Proposed & AugAbEx$^\dagger$ (U, Ex)& \textbf{71.66}& \textbf{47.26}& \textbf{73.05}& \textbf{67.79}& \textbf{37.06}& \textbf{69.85}\\ [0.15cm]
  & AugAbEx$^{*}$ (U, Ex) & 71.37& 46.91& 72.83& 67.67& 36.89& 69.67\\
  \bottomrule
\end{tabular}
\end{subtable}

\vspace{0.5cm}

\begin{subtable}{0.8\textwidth}
\subcaption{Comparison  of AugAbEx method with state-of-the-art approaches}
\label{tab:b-comparison-with-sota}
\begin{tabular}{llcccc}\toprule
 & Approach  &Sample& \multicolumn{3}{c}{BillSum}\\ \midrule
  & && R-1& R-2 & R-L \\ \midrule
         State-of-the-art&Athena (S, Ab)  &100& 51.59 & 29.36& 35.04 \\
  &\citep{2023-hybrid-align-moro} && & &  \\ 
          &DCESumm (U, Ex) &3269&  42.00&  24.28&  38.87  \\
  &\citep{2024-hybrid-DCESumm} && & & \\ 
          &SENDE (S, Ex) &3269&  45.86&  27.52&  31.96 \\ 
  &\citep{2026-soft-labeling-sende} && & &  \\ [0.13cm]
Proposed & AugAbEx (U, Ex) &100& \textbf{55.79} & 31.15&53.50\\
         &  &3269& 55.35& \textbf{31.93}&\textbf{54.27}\\ 
        & &22,218& 55.19& 31.77 & 54.08 \\ 
        \bottomrule
    \end{tabular}
\end{subtable}

\end{table*}

\begin{enumerate}[label=\roman*)]
    \item \textit{Baseline comparison}: Table~\ref{tab:a-comparison-with-baseline} reports the ROUGE scores for the extractive summaries produced by AugAbEx pipeline and two competing baselines on IN-Abs and UK-Abs datasets. First two rows correspond to the  results reported in the original paper over the test sets (size 100)\footnote{Datasets  available at \url{https://github.com/Law-AI/summarization}} of the datasets. The third row displays the ROUGE scores for the TES summaries for the same test set. The quantum jump in the TES summary scores is clearly evident, highlighting that the direction from OAG summaries  begets  quality to the transformed extractive silver summaries, making  the proposed pipeline effective. The performance improvement of more than 40\% is observed for IN-Abs and UK-Abs silver summaries produced by AugAbEx pipeline. The last row shows the macro-averaged summary scores for the full corpus, which are  marginally lower than those on the test set. The slim margin between the TES scores over the test set and full corpus establishes soundness and stability of the proposed AugAbEx pipeline.  
    
    \textit{Though, LexRank and DSDR are established and competent extractive algorithms, guidance from the  experts' opinion present in OAG summaries  significantly improves  the quality of  extractive silver summaries (TES) produced by the proposed pipeline.   } 
    \item \textit{State-of-the-art comparison}: Table~\ref{tab:b-comparison-with-sota} shows comparative  scores of the TES summaries and those generated by three SOTA methods for case summarization. All four methods  use  salient sentences from the case document guided by OAG as input, but adopt different  strategies to craft the final summary. Note that the results of the SOTA methods are reported in the respective original papers on test sets of different sizes.  We match the test sets\footnote{Athena results reported on first 100 documents in the training set. The results for DCESumm and SENDE algorithms  are reported on test set of size 3269 available at \url{https://huggingface.co/datasets} } corresponding to each competing SOTA method and report results accordingly.   
    
    It is clearly evident from the  results that AugAbEx pipeline consistently achieves higher ROUGE F-scores for BillSum dataset against all competing methods. The TES summaries achieve improvement of approximately $8-52\%$ in R-1, R-2, and R-L F-scores, for the test set of 100 documents. Similarly, the performance gain of approximately $16-39\%$ is observed for the test set of $3269$ documents. The improvement achieved by the TES summaries underscores the effectiveness of simple and greedy MMR-based extractive summarizer. AugAbEx pipeline results are highly consistent for all sample sizes, which reinforces the soundness and stability of the proposed method.
    
     \textit{The  scores macro-averaged over the full corpus of approximately $22K$ documents  highlight the role of candidate sentence selection and final summary generation by MMR algorithm, in  preservation of experts' insights in TES summaries.}
\end{enumerate}
%
 \section{Conclusion}
We propose, AugAbEx, an unsupervised two-step pipeline for transforming original abstractive gold summaries (OAG) summaries to extractive silver (TES) summaries. In the first step, the proposed method selects top-k similar sentences from the case document corresponding to each OAG sentence. Subsequently, it leverages a relevance and diversity based method, MMR, as an extractive summarizer for final summary generation. To assess the fidelity of the TES summaries to both OAG summaries and case documents, we propose a comprehensive evaluation framework that spans along structural, lexical, semantic, and domain-specific dimensions.  

Using the proposed framework, we perform extensive experimentation using seven English case summarization datasets. We also conduct Bradley-Terry test, which is a robust statistical procedure for pairwise comparative analysis of two texts. The  experimental results reveal that the extractive silver summaries retain structural, lexical, semantic and  domain-specific similarity with the original abstractive summaries. The results indicate that the proposed pipeline outperforms recent baseline and state-of-the-art summarization approaches, while maintaining explainability. Overall, the proposed AugAbEx pipeline provides an efficient and scalable solution for transforming abstractive legal summaries into high-quality extractive counterparts and can be extended to other domains, languages, and long-document summarization tasks.

\backmatter
\bmhead{Supplementary information}
Not Applicable



\bmhead{Acknowledgements}
We sincerely appreciate Sagar Rathore for his valuable efforts in extracting legal entities from the summaries.

\section*{Declarations}


\begin{itemize}
\item Funding - Not applicable
\item Conflict of interest/Competing interests (check journal-specific guidelines for which heading to use) - Not applicable
\item Ethics approval and consent to participate - Not applicable
\item Consent for publication - Not applicable
\item Data availability - Yes
\item Materials availability - Yes
\item Code availability - Yes
\item Author contribution : 
\end{itemize}
\newpage
\begin{appendices} 
\renewcommand{\thefigure}{\arabic{figure}} 
\setcounter{figure}{7}
\renewcommand{\thetable}{\arabic{table}}  
\setcounter{table}{10}
\section{LegalNER}
\label{appendix-lent}
LegalNER \citep{2022-legal-ner} is a transformer-based model (Roberta + Transition-based Parser) tailored for Indian legal documents using SpaCY framework.  The model is trained to identify $14$ different legal entities prevalent in Indian judgments. A typical Indian judgment has a \textit{Preamble} containing the metadata and the body that has the judgment text. The LegalNER model draws entities from preamble and the judgment. The following Table \ref{tab:legalNer-labels}, borrowed from the original paper, shows the complete list of entities recognized by the LegalNER model in Indian legal judgments. The middle column shows the source portion of the judgment from where the entity is extracted.
\begin{table}[b]
\caption{Definitions of Legal Named Entities}
\label{tab:legalNer-labels}
\centering
\begin{tabular}{@{}llp{7cm}@{}} 
    \toprule
     \textbf{Named Entity}&  \textbf{Extract From} & \textbf{Description} \\  \midrule
     COURT&  Preamble, Judgment& Name of the court which has delivered the current judgment if extracted from the preamble. Name of any court mentioned if extracted from judgment sentences.\\  \midrule 
     PETITIONER&  Preamble, Judgment& Name of the petitioners/appellants/revisionist from current case. \\  \midrule
     RESPONDENT&  Preamble, Judgment& Name of the respondents/defendants/opposition from current case.\\  \midrule
     JUDGE&  Preamble, Judgment& Name of the judges from the current case if extracted from the preamble. Name of the judges of the current as well as previous cases if extracted from judgment sentences. \\  \midrule
     LAWYER&  Preamble& Name of the lawyers from both the parties. \\  \midrule
     DATE&  Judgment& Any date mentioned in the judgment. \\  \midrule
     ORG&  Judgment& Name of organizations mentioned in text apart from the court.\\  \midrule
     GPE&  Judgment& Geopolitical locations which include names of states, cities, villages. \\  \midrule
    STATUTE& Judgment&Name of the act or law mentioned in the judgment. \\  \midrule
    PROVISION& Judgment&Sections, sub-sections, articles, orders, rules under a statute. \\  \midrule
    PRECEDENT& Judgment&All the past court cases referred to in the judgment as precedent. Precedent consists of party names + citation(optional) or case number (optional). \\  \midrule
    CASE\_NUMBER& Judgment&All the other case numbers mentioned in the judgment (apart from precedent) where party names and citation is not provided. \\  \midrule
    WITNESS& Judgment&Name of witnesses in current judgment. \\  \midrule
     OTHER\_PERSON&  Judgment& Name of all the persons that are not included in petitioner, respondent, judge and witness.\\
     \bottomrule
\end{tabular}
\end{table}
\begin{figure*}[h]
    \centering
    \includegraphics[width=\textwidth]{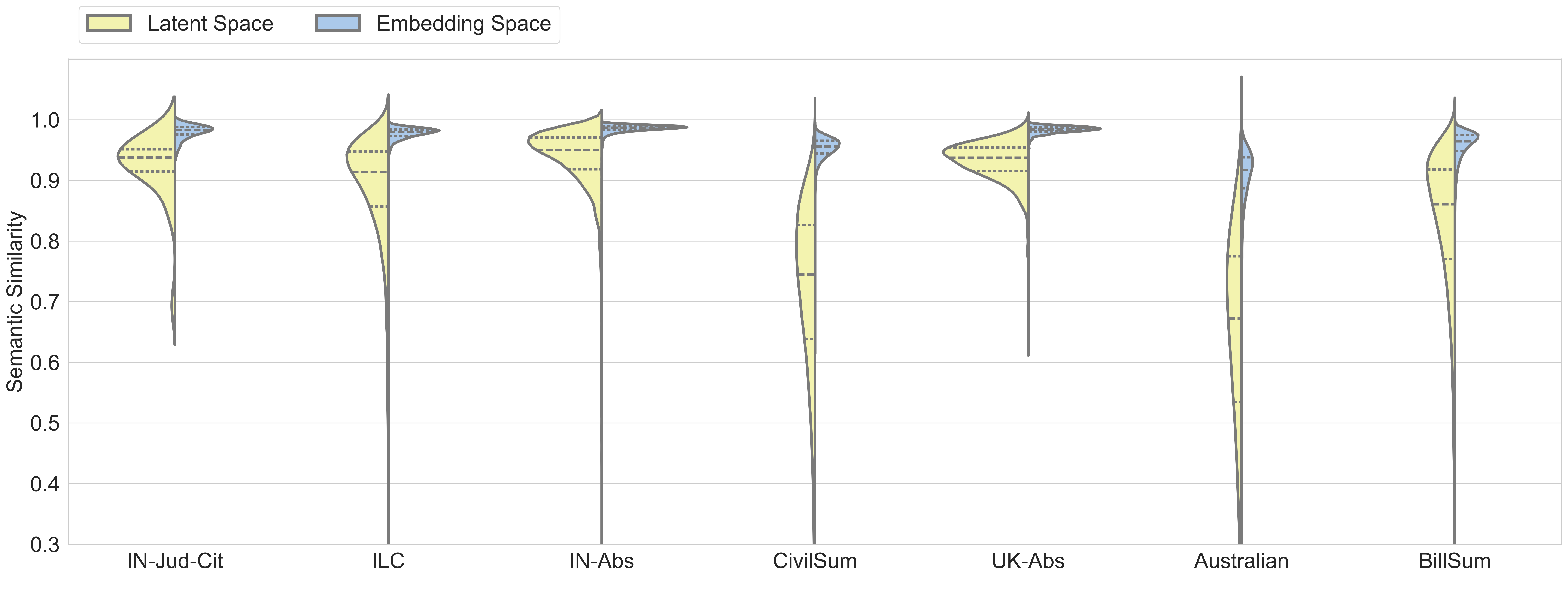} 
    \caption{Comparison between the distribution of similarity scores of OAG and TES summaries in the latent and embedding space.}
    \label{fig:distribution-plot-ss-cd}
\end{figure*}
\section{Distribution of scores for Semantic Similarity}
\label{a:contrast-latent-embed}
Figure \ref{fig:distribution-plot-ss-cd} juxtaposes the distributions of semantic similarity scores of OAG and TES summaries in the LSA latent space and the LegalBert embedding space. It is apparent that the semantic similarity between the two summaries is higher in the LegalBert embedding space compared to the LSA latent space for all datasets. This clearly shows that LegalBert is able to capture semantic relationships between legal documents more effectively.
\begin{figure*}[h]
    \centering
    \includegraphics[width=0.8\textwidth]{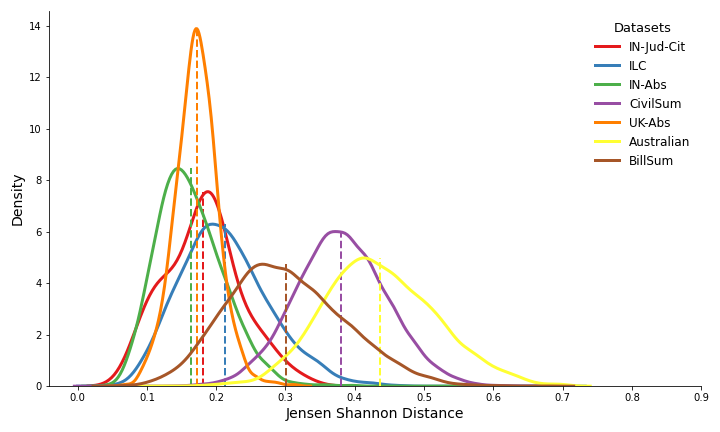} 
    \caption{Density plot of Jensen-Shannon distances between the probability distributions of terms in OAG and TES summaries for all datasets.}
    \label{fig:jsd-terms-density-plot-all-datasets}
\end{figure*}
\section{Density plot of the Jensen-Shannon Distance for all Datasets}
\label{a:jsd-all-density}
Figure \ref{fig:jsd-terms-density-plot-all-datasets} shows the distributions of Jensen-Shannon distances between OAG and TES summaries for the seven datasets. Relative positions of the density curves on the X-axis provide insights into the level of content similarity between OAG and TES summaries for each dataset. Lower values for IN-Jud-Cit, ILC, IN-Abs and UK-Abs datasets (Table~\ref{tab:Rouge-JSD-OAG-TES}), and higher peaks suggest greater alignment in the information captured by the TES summaries. Conversely, broader distributions  and flatter peaks, farther right on the X-axis for CivilSum, Australian and BillSum datasets indicate more divergence between the two summaries at the lexical level.
\section{Original abstractive summary structured as phrases}\label{a:example-phrasal}
We present one sample summary, each from CivilSum\footnote{Dataset downloaded from \url{https://github.com/ra-MANUJ-an/CivilSum?tab=readme-ov-file}} and Australian\footnote{\url{https://archive.ics.uci.edu/dataset/239/legal+case+reports}} dataset, to display the phrasal structure. The documents were chosen randomly from the datasets based on the criteria of summary length. Summary for the CivilSum judgment has short phrases, a characteristic which manifests as low values of  structural and lexical metrics. However, it contains a few references to legal entities.  The TES summary for Australian case document similarly exhibits lower values for structural and lexical similarities with the OAG summary shown here. Note that it does not reference statutes and provisions, which is reflected in the dataset's low \textit{Lent-cnt} in both OAG and TES summaries (Table \ref{tab:bt-score-Legal-Ner-entity}).  
\begin{tcolorbox}[title=Abstractive Gold-Standard Summary for CivilSum Case Document - ID 22413, colback=yellow!10!white, colframe=black, boxrule=0.5mm, sharp corners=south]
Constitution of India, Articles 14, 16 and 21 - Central Services (Medical Attendance) Rules, 1944, Rule 1 Note 2(iv) - Medical bills - Reimbursement of - Serving Govt. servants and retired Govt. servants are to be classified separately under Art. 14, but not under Art. 21 - Article 21 provides a constitutional obligation of the Government to provide medical facility to retired as well as in service Govt. servants - Allocation of limited funds needs to be made for the Scheme - Prior approval of medical board is not required for emergency treatments - The Central Government Health Scheme is being gradually extended and its unavailability in some areas does not make it discriminatory - Administrative Tribunals Act, 1985, Section 19 empowers the Tribunal to issue mandamus.
[Paras 14-19, 24, 25, 26, 27(a-c)]
\end{tcolorbox}
\vspace{0.3cm}
\begin{tcolorbox}[title=Gold Standard Summary for Australian Case Document - ID  06\_1, colback=yellow!10!white, colframe=black, boxrule=0.5mm, sharp corners=south]
practice and procedure - application for leave to appeal - authorisation of multiple infringements of copyright established - prior sale of realty of one respondent to primary proceedings - payment of substantial part of proceeds of sale to offshore company in purported repayment of loan - absence of material establishing original making and purpose of loan - mareva and ancillary orders made by primary judge - affidavits disclosing assets sworn - orders made requiring filing of further affidavits of disclosure and cross-examination of one respondent to primary proceedings on her disclosure affidavit - no error in making further ancillary orders - leave refused
\end{tcolorbox}





\end{appendices}

\clearpage
\bibliography{sn-bibliography}
\end{document}